\documentclass[aps,prx,twocolumn,floatfix,10pt,showpacs,superscriptaddress,longbibliography]{revtex4-1}
\usepackage[bookmarks=true,linkcolor=blue,urlcolor=blue,colorlinks,citecolor=blue]{hyperref}
\usepackage{graphicx}
\usepackage{epstopdf}
\usepackage{dcolumn}
\usepackage{bm}
\usepackage{amsmath}
\usepackage{amssymb}
\usepackage{latexsym}
\usepackage{epsfig}
\usepackage{amsbsy}
\usepackage{array}
\usepackage{amssymb}
\usepackage{setspace}
\usepackage{subfigure}
\usepackage{caption}
\usepackage{natbib}
\usepackage{multirow}
\renewcommand{\raggedright}{\leftskip=0pt \rightskip=0pt plus 0cm}
\usepackage{hyperref}
\usepackage{float}
\usepackage{enumerate}
\UseRawInputEncoding

\begin{document}
\bibliographystyle{apsrev4-1-title}

\graphicspath{{figures/}}

\title{Learning Order Parameters from Videos of Dynamical \\Phases for Skyrmions with Neural Networks}

\author{Weidi Wang}
\affiliation{Department of Physics, Beijing Normal University, Beijing 100875, China}
\affiliation{The Center for Advanced Quantum Studies, Beijing Normal University, Beijing 100875, China}

\author{Zeyuan Wang}
\affiliation{Department of Physics, Drexel University, Philadelphia, PA 19104, USA}

\author{Yinghui Zhang}
\affiliation{College of Education for the Future, Beijing Normal University, Beijing 100875, China}

\author{Bo Sun}
\email[Bo Sun: ]{tosunbo@bnu.edu.cn}
\affiliation{College of Education for the Future, Beijing Normal University, Beijing 100875, China}
\affiliation{School of Artificial Intelligence, Beijing Normal University, Beijing 100875, China}

\author{Ke Xia}
\email[Ke Xia: ]{kexia@bnu.edu.cn}
\affiliation{The Center for Advanced Quantum Studies, Beijing Normal University, Beijing 100875, China}
\affiliation{Beijing Computational Science Research Center, Beijing 100193, China}

\date{\today}

\begin{abstract}
The ability to recognize dynamical phenomena (e.g., dynamical phases) and dynamical processes in physical events from videos, then to abstract physical concepts and reveal physical laws, lies at the core of human intelligence.
The main purposes of this paper are to use neural networks for classifying the dynamical phases of some videos and to demonstrate that neural networks can learn physical concepts from them.
To this end, we employ multiple neural networks to recognize the static phases (image format) and dynamical phases (video format) of a particle-based skyrmion model.
Our results show that neural networks, without any prior knowledge, can not only correctly classify these phases, but also predict the phase boundaries which agree with those obtained by simulation.
We further propose a parameter visualization scheme to interpret what neural networks have learned.
We show that neural networks can learn two order parameters from videos of dynamical phases and predict the critical values of two order parameters.
Finally, we demonstrate that only two order parameters are needed to identify videos of skyrmion dynamical phases.
It shows that this parameter visualization scheme can be used to determine how many order parameters are needed to fully recognize the input phases.
Our work sheds light on the future use of neural networks in discovering new physical concepts and revealing unknown yet physical laws from videos.
\end{abstract}

\maketitle

\section{Introduction}
Humans, even young infants, can recognize and classify objects and their motions, and interpret what has happened in them, deduce (to some extent) what should happen, and imagine what would happen in other situations \cite{yi2019clevrer}.
This reflects the ability of human beings to reason logically based on physical events and then to extract experience and laws (or patterns) from them.
Therefore, we can summarize the process of human discovering of physical laws as (i) identifying and classifying physical events, (ii) abstracting physical concepts, (iii) revealing physical laws.
With the rapid development of artificial intelligence, it has become possible to use machine learning to perform the first two processes.

Machine learning has extremely broad applications in many fields and has achieved breakthrough progress with bright prospects \cite{jordan2015machine}. Previous applications of machine learning techniques in condensed matter physics centered on representing quantum states \cite{carleo2017solving}, analyzing experimental data \cite{zhang2019machine,rem2019identifying}, classifying different phases and their transitions \cite{wang2016discovering,carrasquilla2017machine,van2017learning} and discovering physical concepts \cite{iten2020discovering}.

One of the earliest machine learning algorithms was the artificial neural network, a computational model designed to simulate biological learning \cite{goodfellow2016deep}.
From the perspective of network architecture, there are three fundamental neural networks: (i) feedforward neural network (FNN), that can fit the nonlinear relationship between any two columnar data \cite{haykin1994neural}, (ii) convolutional neural network (CNN), which is suitable for image data recognition and classification, and (iii) recurrent neural network (RNN), that is applicable to processing sequential data.
In Ref.~\cite{patterson2017deep}, the authors hold the belief that the input data type should match the used network architecture.
In other words, when the input is columnar data, image data, or processing sequential data, the corresponding network architecture is FNN, CNN, or RNN [specifically Long Short-Term Memory (LSTM)], respectively. While when the input is video data, one should apply the hybrid CNN and LSTM architecture.

The type of input data reflects the complexity of the problem under study, and the application of machine learning techniques in the field of phase transitions recognition ranges from simple to complex.
Initially, machine learning was applied to classify phases and phase transitions that are primarily static, so the corresponding methods were generally FNN \cite{iakovlev2018supervised,iakovlev2019profile,
venderley2018machine} and CNN \cite{araki2019phase,beach2018machine,
park2017classification}.
In the last two years, RNN started being used for studying processing sequential data, mainly including the identification of dynamical phases \cite{van2018learning} and dynamical processes \cite{deviatov2019recurrent}, as well as the prediction of qubit dynamics \cite{gupta2018machine,flurin2020using}.
However, the videos of dynamical phases or processes have not been considered so far.
There are two main reasons for this situation.
The first is that, compared with sequential data, the video data has one additional dimension and a larger volume, thus contains more information and is more difficult to identify and work with.
Secondly, the identification of dynamical phases and processes usually uses supervised learning, and this procedure needs to label the training data manually, which requires a lot of time and cost.
Moreover, previous work \cite{van2018learning,deviatov2019recurrent,
gupta2018machine,flurin2020using} only regarded neural networks as a black box, profound analysis of it being outside of the interest of its authors.
But it is essential for the future discovery of physical laws to use neural networks for classifying videos of dynamical phases and to analyze what they have learned.

In this paper we consider the dynamics of skyrmions based on a particle model, since many kinds of systems can be adequately modeled as collectively interacting particles moving in quenched disorder, with the transition from a pinned phase to a sliding phase under an applied drive \cite{reichhardt2015collective}.
Examples include colloids on various types of substrates \cite{bohlein2012observation}, Wigner crystals and vortices in type-II superconductors \cite{fily2010critical}.
In the following, we first use a variety of neural networks to classify static and dynamical phases generated by this model and use the trained neural networks to predict the phase boundaries.
Then we analyze the neural networks by feature map visualization and a parameter visualization scheme.
The latter is based on the idea of representation learning \cite{bengio2013representation}, combined with variational autoencoders \cite{kingma2013auto} and supervised learning. This represents the main achievement of our work.
Last but not least we explore whether the network can determine the number of order parameters by changing the number of neurons, and the influence of using an unevenly distributed training set on the parameter visualization scheme.

\section{Model}\label{model}
We consider a particle model for skyrmions in metallic chiral magnets which can describe the interaction between skyrmions and random disorder \cite{PhysRevB.87.214419}. The equation of motion for a single skyrmion $i$ is \cite{reichhardt2015collective}:
\begin{equation}
\alpha_d \bm{v}_i=\bm{F}_M+\bm{F}^{ss}_i+\bm{F}^{sp}_i+\bm{F}_D. \label{skyrmion}
\end{equation}
Here $\bm{v}_i$ means the skyrmion velocity. The damping term $\alpha_d$ adjusts the velocity to the net force acting on the skyrmion, while the first term on the right-hand side is the Magnus term $\bm{F}_M=-\alpha_m\hat{z}\times\bm{v}_i$ that turns the skyrmion velocity into the perpendicular direction. According to Ref.~\cite{reichhardt2015collective}, we impose the constraint $\alpha_d^2 + \alpha_m^2 = 1$ and take $\alpha_m/\alpha_d=9.962$. The skyrmion-skyrmion interaction force is $\bm{F}^{ss}_i=F^{s0}\sum_{j=1}^{N_s} \hat{\bm{r}}_{ij} K_1(R_{ij}/\xi_s)$, where $F^{s0}=1$, $R_{ij}=\left|\bm{r}_i-\bm{r}_j\right|$, $\hat{\bm{r}}_{ij}=(\boldsymbol{r}_i-\boldsymbol{r}_j)/R_{ij}$, $K_1$ is the modified Bessel function and $\xi_s=3$ represents the size of skyrmions. The pinning force is $\bm{F}^{sp}_i=\sum_{j=1}^{N_p}J_d\exp(-\bm{r}_d/\xi_d)$, where $J_d$ characterizes the strength of the pinning, $\bm{r}_d$ represents the radial vector between the $i$th skyrmion and the $j$th pinning center, and $\xi_d=0.3$ means the size of defects. The pinning force $\bm{F}^{sp}$ is generated by a randomly placed, non-overlapping harmonic trap, and its maximum pinning force is $\bm{F}_p$ \cite{reichhardt2015collective}. The driving term is $\bm{F}_D=2\pi\hbar e^{-1}\hat{z}\times\bm{J}_B$, where $\bm{J}_B$ is the externally applied current. The system size is $L\times L$ with $L=36$, while the density of skyrmions ($\rho_s$) and pinning sites ($\rho_p$) are $0.1$ and $0.3$, respectively. In simulations, we use periodic boundary conditions in the $x$ and $y$ directions. In order to express the various interaction forces more concisely, we set the unit force to $F_0\sim10^{-5}$ N/m and the unit length to $l_0\sim10$ nm \cite{reichhardt2015collective}.

\begin{figure}
	\centering
	\scalebox{0.5}[0.5]{\includegraphics[width=0.95\textwidth]{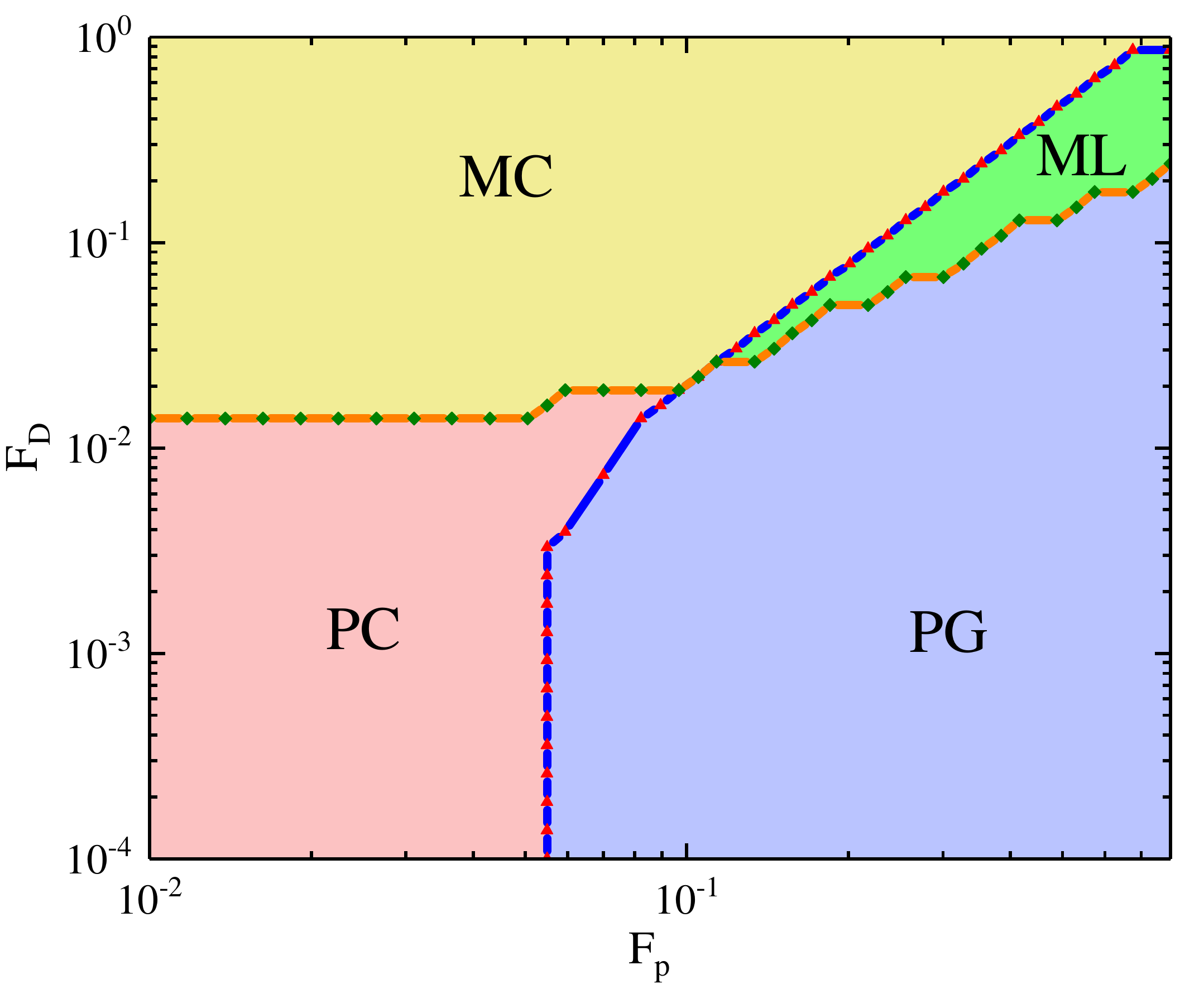}}
	\caption{\raggedright The dynamical phase diagram for skyrmions as $\bm{F}_p$ and $\bm{F}_D$ vary. MC: moving crystal; ML: moving liquid; PG: pinned amorphous glass; PC: pinned crystal. The solid blue line is the phase boundary between crystal and amorphous states, and the solid orange line is the phase boundary between moving and pinned states. We used a grid of $840$ points on the square lattice to construct the dynamical phase diagram. Since the initial positions of skyrmions and defects are random, possibly affecting thus the formed phase diagram, the value for each point is averaged over $10$ simulations. \hfill
    \label{fig1}}
\end{figure}

The numerical details of Eq.~\eqref{skyrmion} are shown in Ref.~\cite{PhysRevB.87.214419}. In each simulation, we calculate $4000$ iterations and draw a figure of skyrmions distribution for each iteration.
In order to make the motion information more obvious, we take an image every 15 iterations, and thus we obtain almost $270$ images. Because a standard video has 30 frames per second, we finally produce thus a nine-second video during each calculation.
We can get different $\bm{F}_D$ and $\bm{F}_p$ by changing the magnitude of the pinning strength $J_d$ and the applied current $\bm{J}_B$, and then obtain four different dynamical phases in video format (see Supplemental Material \footnote{See Supplemental Material at url for videos of the four dynamical phases and further detailed information.}).
The diagram of four dynamical phases is shown in Fig.~\ref{fig1}. When $\bm{F}_p<0.055$, the skyrmions form a pinned crystal state (PC), which transforms into a moving crystal state (MC) as the driving force $\bm{F}_D$ increases. When $\bm{F}_p>0.055$, we first find a pinned amorphous glass state (PG) at low drives. As the driving force $\bm{F}_D$ increases, it converts to moving liquid (ML) later and then to moving crystal. From the diagram and videos of the four dynamical phases, it is easy to deduce that the driving term $\bm{F}_D$ determines the velocity of skyrmions movement, while the strength of the interaction between skyrmion and skyrmion $\xi_s$ and the one between skyrmion and defect $\xi_d$ determine whether it forms a crystal or an amorphous state.

\section{Classification of static and dynamical skyrmion phases}\label{sec.3}
We study the classification of videos of four dynamical phases for the particle-based skyrmion model.
Our goal is to understand whether supervised learning with neural networks can classify these phases exactly.
In particular, we also hope that the neural network can accurately predict the phase boundary between them.

Image recognition is the basis of video recognition, so we first use the CNN to recognize the images of static skyrmion crystal and amorphous phases.
Besides, we find that the problem of recognizing videos of dynamical phases recognition is similar to the problem of video action recognition in machine learning.
In Ref.~\cite{carreira2017quo}, the authors summarize five machine learning methods of video action recognition: (i) CNN-LSTM, (ii) three dimensional convolutional neural network (3D-CNN), (iii) two-stream, (iv) 3D-fused two-stream, (v) two-stream 3D-CNN.
Here we use the first two basic methods (CNN-LSTM and 3D-CNN) to identify the dynamical phases of skyrmions.

Since our machine learning method belongs to supervised learning, we also need to label the training image data and video data.
From videos of the four dynamical skyrmion phases one can observe that crystal and amorphous states are divided by spatial information, moving and pinned states being moreover separated by temporal information.
In order to classify crystal and amorphous states, we calculate the radial distribution function (RDF) and the standard deviation of radial distribution function (SDRDF), whose value is denoted by $S$.
We also define mean-velocity ($\bar{v}$) to categorize moving and pinned states (the corresponding expressions are shown in Appendix~\ref{app.A}).
Apparently, we can consider SDRDF as the spatial order parameter, and we can consider mean-velocity as the temporal order parameter.

\subsection{Classification of static skyrmion crystal and amorphous phases by the CNN}\label{sec.3a}
\begin{figure}[tb]
	\centering
	\scalebox{0.5}[0.5]{\includegraphics[width=0.97\textwidth]{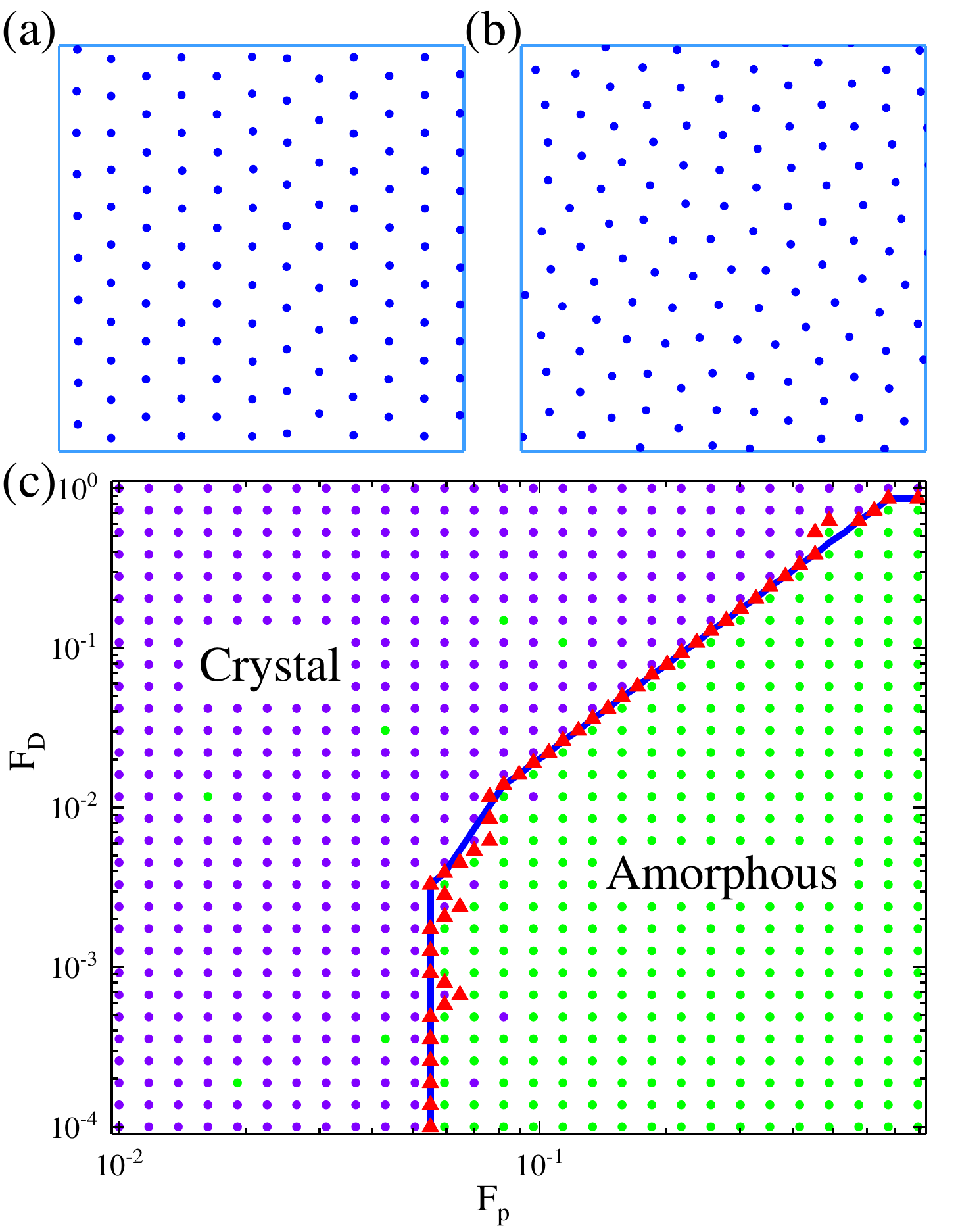}}
	\caption{\raggedright Examples of (a) crystal and (b) amorphous states. (c) The phase diagram of crystal and amorphous states predicted by the CNN. The blue line is the phase boundary between crystal and amorphous phase, the same as that in Fig.~\ref{fig1}, so the points on the left of the blue line represent the data for crystals and the points on the right of the blue line represent the data for amorphous states. After these test data are sent to the CNN, this will give the predicted labels. The color of the dots represents the predicted label (i.e., purple represents crystal, green represents amorphous state). The red triangles represent the phase boundary predicted by CNN. The testing accuracy is $98.33\%$ and $14$ test data were misclassified. These misclassified data are green points in the crystal area, purple points in the amorphous area, and the points between the predicted phase boundary and the simulated phase boundary.\hfill
    \label{fig:CNN}}
\end{figure}

We first train the CNN to distinguish the crystal phase with sixfold ordering  [Fig.~\ref{fig:CNN}(a)] from the unordered amorphous phase [Fig.~\ref{fig:CNN}(b)].
The $16800$ labeled images of crystal and amorphous phases are generated through simulations at various interaction strengths ($\xi_d$ and $\xi_s$), in which $80\%$ of the data are for training and $20\%$ for validation.

The CNN starts with a convolutional layer consisting of $32$ filters of size $5\times5$ with rectified linear (ReLu) activation functions. The output of this layer is sent to the batch normlization (BN) layer before applying $2\times2$ max pooling.
To extract features in a sufficient manner the network is then fed into the same convolutional layer, BN layer, and into max pooling layers with ReLu activation.
After that, a fully-connected layer with $1024$ ReLu units is followed before the ultimately softmax output layer with two neurons.
The actual training of the network is completed by minimizing the cross entropy by using the Adam optimizer.
Finally, we save the model parameters (weights and biases) when the training is completed and the validation accuracy reaches the best result.
Considering the randomness of neural networks for each training, we train five identical networks at the same time and then select the optimal model parameter to test according to the highest validation accuracy.

We wish to know whether the CNN can predict the phase boundary of skyrmion crystal and amorphous states, so we evenly select 840 images in the phase diagram of Fig.~\ref{fig1} as a test set to see the performance of the CNN on it.
From Fig.~\ref{fig:CNN}(c), we can see that the trained CNN can successfully recognize the skyrmion crystal and amorphous states in the test set and can predict the phase boundary (red triangles) which is in good agreement with that obtained by calculating SDRDF (blue solid line).
The good performance of the CNN on the test set shows that the CNN can capture spatial information to distinguish between crystal and amorphous states.

\subsection{Classification of four dynamical skyrmion phases by the CNN-LSTM}
\label{sec.3b}
\begin{figure}[tb]
	\centering
	\scalebox{0.5}[0.5]{\includegraphics[width=1\textwidth]{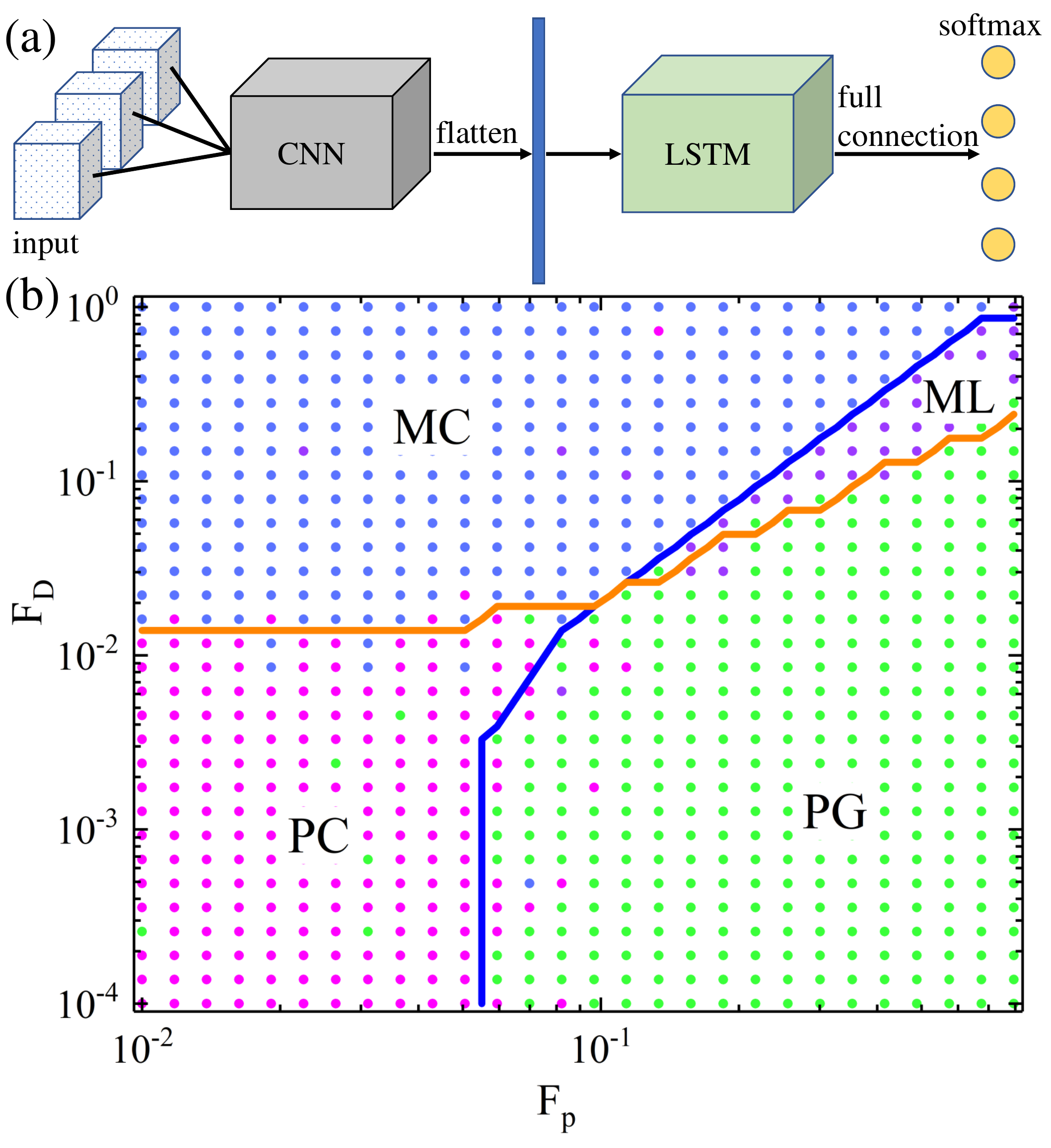}}
	\caption{\raggedright (a) The structure of the hybrid CNN and LSTM. (b) The phase diagram predicted by the CNN-LSTM. The blue and orange lines are the phase boundaries between crystal and amorphous phases, moving and pinned phases, respectively, which are the same as those in Fig.~\ref{fig1}. After these test data are sent to the CNN-LSTM, the network will give the prediction of labels. The color of the dots represents the predicted label (i.e., purple represents ML, green represents PG, magenta represents PC, and blue represents MC). The testing accuracy is $94.40\%$ and $47$ test data were misclassified. These misclassified data are the points with different colors in each area, such as the green points in the PC area. \hfill
    \label{fig:CNN-LSTM}}
\end{figure}

In this section, we show a universal architecture integrating CNN and LSTM that can recognize the four dynamical phases of skyrmions in video format.
In the hybrid CNN and LSTM architecture, the CNN is specially designed to extract spatial features, while the LSTM is used to extract temporal features.
Then temporal and spatial features are fused by the softmax function to achieve the purpose of video recognition.
Each input data is a video composed of $10$ frames of $36\times36$ images which are the skyrmions distribution figures drawn by the results of each iteration during the simulation.
We train the CNN-LSTM using a dataset of about $15,960$ videos, in which $80\%$ for training and $20\%$ for validation.

The CNN-LSTM architecture is shown in Fig.~\ref{fig:CNN-LSTM}(a).
The structure of CNN here is the same as that of CNN in Sec.~\ref{sec.3a}, except that the size of the convolutional layer and max pooling layer is $3\times3$. The output of CNN is flattened to the LSTM consisting of $50$ neurons.
The final layer of our neural network is a softmax layer with four neurons, the output of which being the probabilities of the four dynamical phases.

Additionally we evenly selected $840$ video data in the phase diagram of Fig.~\ref{fig1} as a test set to analyze the performance of the CNN-LSTM on it.
As can be seen from the predicted diagram in Fig.~\ref{fig:CNN-LSTM}(b), the CNN-LSTM is able to classify the four dynamical phases of skyrmions.

\subsection{Classification of four dynamical skyrmion phases by the 3D-CNN} \label{sec.3c}

For the sake of getting the spatial and temporal information, a direct approach to video recognition using deep networks is to combine the convolutional operation with temporal features \cite{herath2017going}.
Now we use the 3D-CNN, which is superior to two-dimensional CNN \cite{ji20123d}, for recognizing videos of four dynamical skyrmion phases.

\begin{figure*}[tb]
\centering
\scalebox{0.9}[0.9]{\includegraphics[width=1\textwidth]{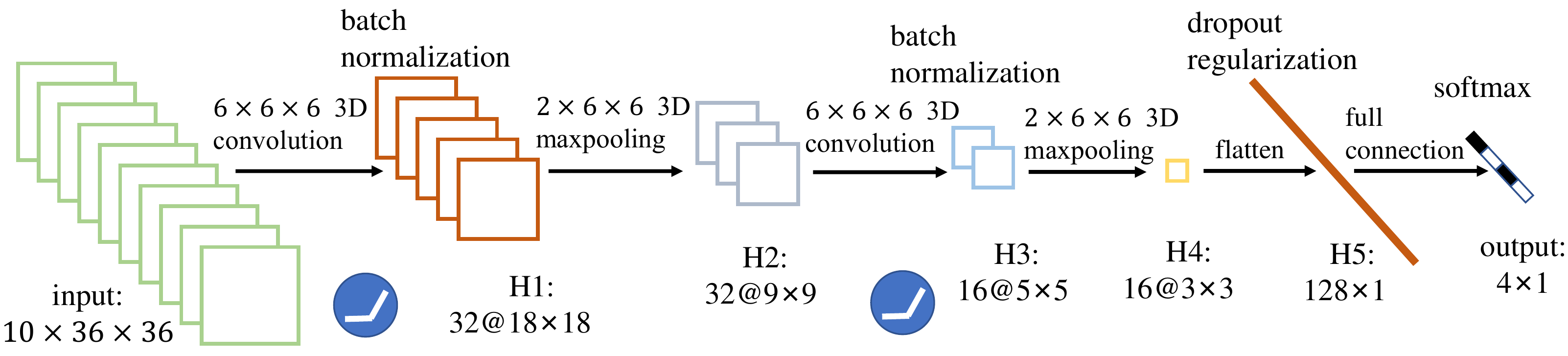}}
\caption{\raggedright The architecture of the 3D-CNN. The bule circle represents ReLu. We use the same batch normalization and 3D max pooling layers. The first and second 3D convolutional layers are also the same except the numbers of convolutional kernels ($32$ and $16$ respectively). \hfill
\label{fig3}}
\end{figure*}

\begin{figure}[tb]
	\scalebox{0.5}[0.5]{\includegraphics[width=0.97\textwidth]{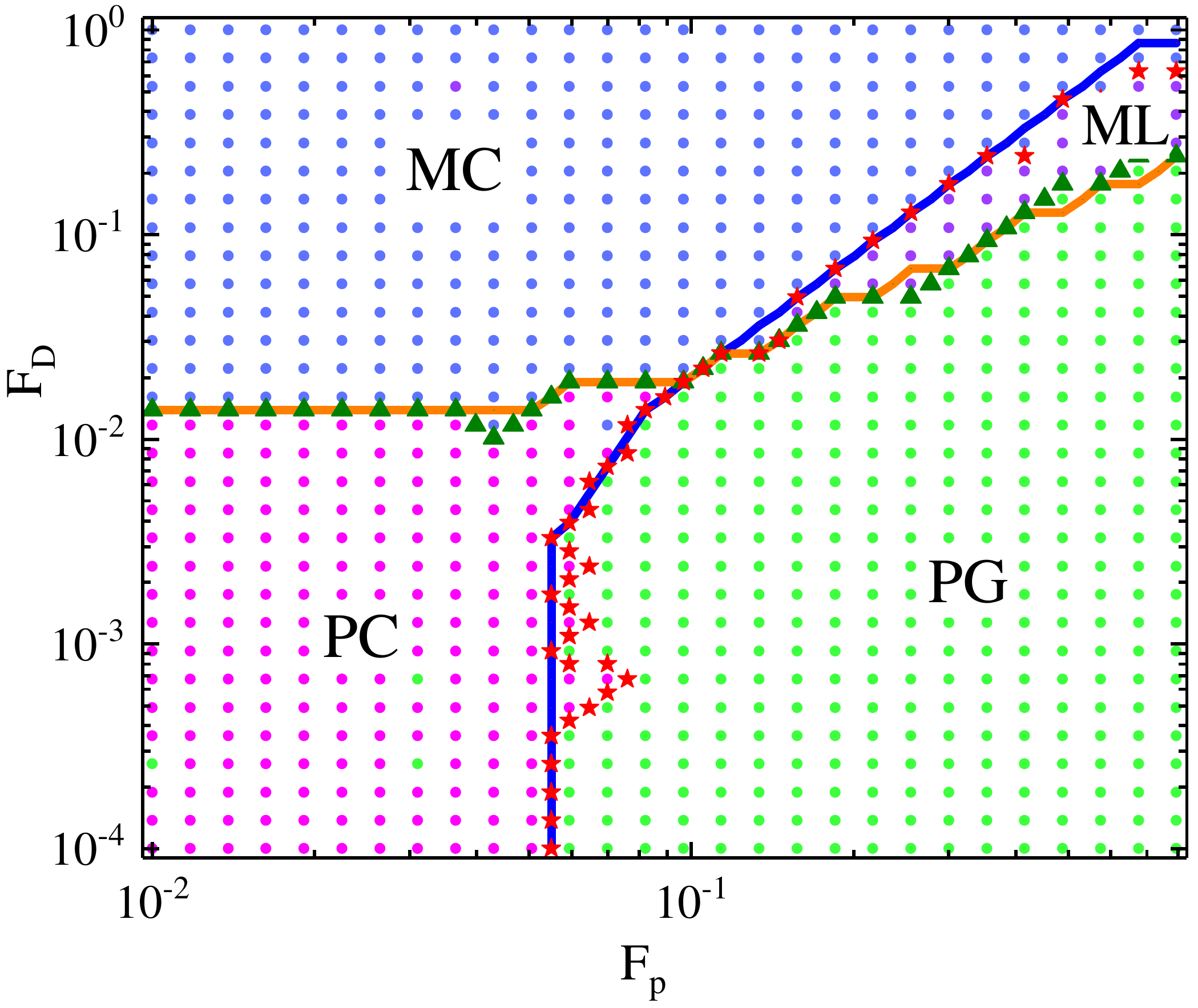}}
	\caption{\raggedright The phase diagram predicted by the 3D-CNN.
The blue and orange lines are the phase boundaries between crystal and amorphous states, moving and pinned states, respectively, which are the same as those in Fig.~\ref{fig1}. After these test data are sent to the 3D-CNN, the network will give the prediction of labels. The color of the dots represents the predicted label (i.e., purple represents ML, green represents PG, magenta represents PC, and blue represents MC). The red pentagram represents the predicted phase boundary between crystal and amorphous phase, while the green triangle represents the predicted phase boundary between moving and pinned phase. The testing accuracy is $97.74\%$ and $19$ test data were misclassified. These misclassified data are the points with different colors in each area (e.g., the green points in the PC area) and the points between the predicted phase boundary and the simulated phase boundary.
 \hfill
		\label{fig:pred3dcnn}}
\end{figure}

The network architecture is shown in Fig.~\ref{fig3}.
The training data and test data are the same as those used by the CNN-LSTM.
After the training is finished, we wish to know whether the 3D-CNN can predict the phase boundary of dynamical phases for skyrmions.
From Fig.~\ref{fig:pred3dcnn} we can see that the trained 3D-CNN can predict the crystal and amorphous phase boundary (red pentagram) which is in good agreement with the phase boundary obtained by simulation (blue solid line).
Moreover, the trained 3D-CNN can predict the moving and pinned phase boundary (green triangle) which is in good agreement with the phase boundary obtained by simulation (orange solid line).
In conclusion, the 3D-CNN can not only classify four dynamical phases of skyrmions but is able to predict the boundaries between them as well.

\section{Analysis of the classification processes}
The results of our previous studies based on machine learning have shown that neural networks can extract spatial and temporal information to recognize static and dynamical phases of skyrmions.
A question which raises naturally is: what did the neural networks learn?
In Ref.~\cite{wang2016discovering}, the author finds the physical concepts (e.g., the order parameter and structure factor) for distinguishing phases by visualizing the samples in the four-dimensional feature space composed of principal components.
By visualizing the weights of the hidden layer to analyze what the neural network has learned, other studies find that the weight of the hidden layer is indeed related to some physical quantities (e.g., the total magnetization and topological charge) \cite{iakovlev2018supervised,suchsland2018parameter}.
This motivates us to perform a similar analysis of our neural networks to explore what they learned.

\subsection{Feature map visualization in the CNN}
Despite the success of CNN in many fields such as computer vision, it has long been viewed as a ``black box'' and it is still unclear why it performs so well \cite{krizhevsky2012imagenet}.
Although it is impossible to open this black box, one can do some visualization investigations.
The studies of visualization can be divided into two main categories. One is feature map visualization, which does not analyze the specific parameters of the convolution kernel, but first selects some images, propagates them once forward in the existing model, and then visualizes the feature map in the convolution kernel.
The other is parameter visualization that focuses on the analysis of parameters in the convolution kernel or neurons.

\begin{figure}[tb]
	\scalebox{0.5}[0.5]{\includegraphics[width=0.98\textwidth]{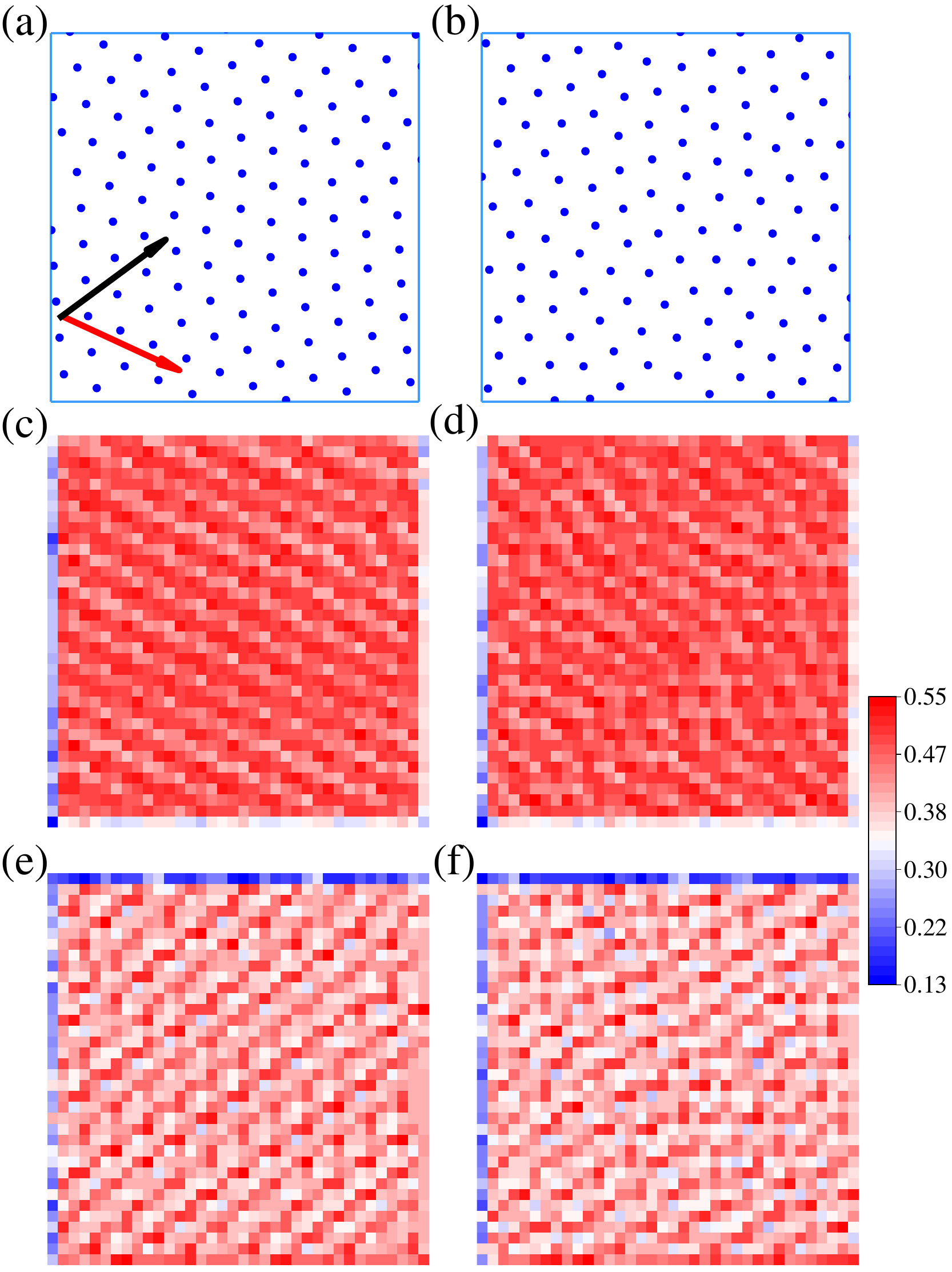}}
	\caption{\raggedright The original image of (a) crystal and (b) amorphous phase. (c) and (e) are feature maps extracted from (a) by CNN. (d) and (f) are feature maps extracted from (b) by CNN.
\hfill
		\label{fig:twofeature}}
\end{figure}

We first visualize the feature maps of the CNN which has recognized the images of skyrmion crystal and amorphous phases.
We select respectively a crystal and  amorphous image as shown in Fig.~\ref{fig:twofeature}(a) and \ref{fig:twofeature}(b).
It can be seen that skyrmions in the crystal are arranged in the direction of black and red arrows, while skyrmions in the crystal are disordered.
To find out whether the neural network can learn the information of crystal orientation, after preprocessing, the images are transferred to the CNN which has been trained before.
Then we visualize the feature maps of the first convolutional layer.
Two feature maps of the crystal and amorphous image are respectively shown in Fig.~\ref{fig:twofeature}.
Comparing Fig.~\ref{fig:twofeature}(c) and \ref{fig:twofeature}(d), we can see that the CNN has extracted the information of oblique downward arrangement [i.e., the direction of the red arrow in Fig.~\ref{fig:twofeature}(a)] from the images.
From the comparison of Fig.~\ref{fig:twofeature}(e) and \ref{fig:twofeature}(f), we can see that the CNN has extracted the information of oblique upward arrangement [i.e., the direction of the black arrow in Fig.~\ref{fig:twofeature}(a)] from the images.
Therefore, we can conclude that the CNN extracts some information about the spatial structure, and recognizes the images of crystal and amorphous phases based on this information.

\subsection{Parameter visualization in the CNN}
The feature map visualization visually shows the information of what CNN learned from the original image.
Now we want to perform a parameter visualization to explore the relationship between the values of neuron outputs in CNN and physical parameters (e.g., the order parameter).
In Ref.~\cite{iten2020discovering}, the authors propose a scheme for visualizing the intermediate parameters of a neural network based on the idea of representation learning.
They consider the former part of the neural network as an encoding process and the latter part as a decoding process.
When the training of a neural network is completed, the information stored by neurons in the middle layer consists of physical quantities that are closely related to the physical problem under study.
For example, in the study of damped pendulum motion, the information preserved by neurons in the middle layer is related to the spring constant and damping coefficient.

With this idea, we propose a scheme that can reveal what the neural networks have learned and we refer to it as a parameter visualization scheme.
Specifically, we add a newly fully-connected layer in front of the original CNN output layer and set the number of neurons to 1 because the classification of skyrmion crystal and amorphous phases from physics requires only one order parameter (i.e., SDRDF).
And the information of input crystal and amorphous images is compressed into one neuron in this fully-connected layer after a series of convolutional and pooling operations.
We are interested to know whether this neuron output is related to SDRDF (the details of parameter visualization scheme are shown in Appendix~\ref{app.B}).

\begin{figure}[b]
	\scalebox{0.5}[0.5]{\includegraphics[width=0.92\textwidth]{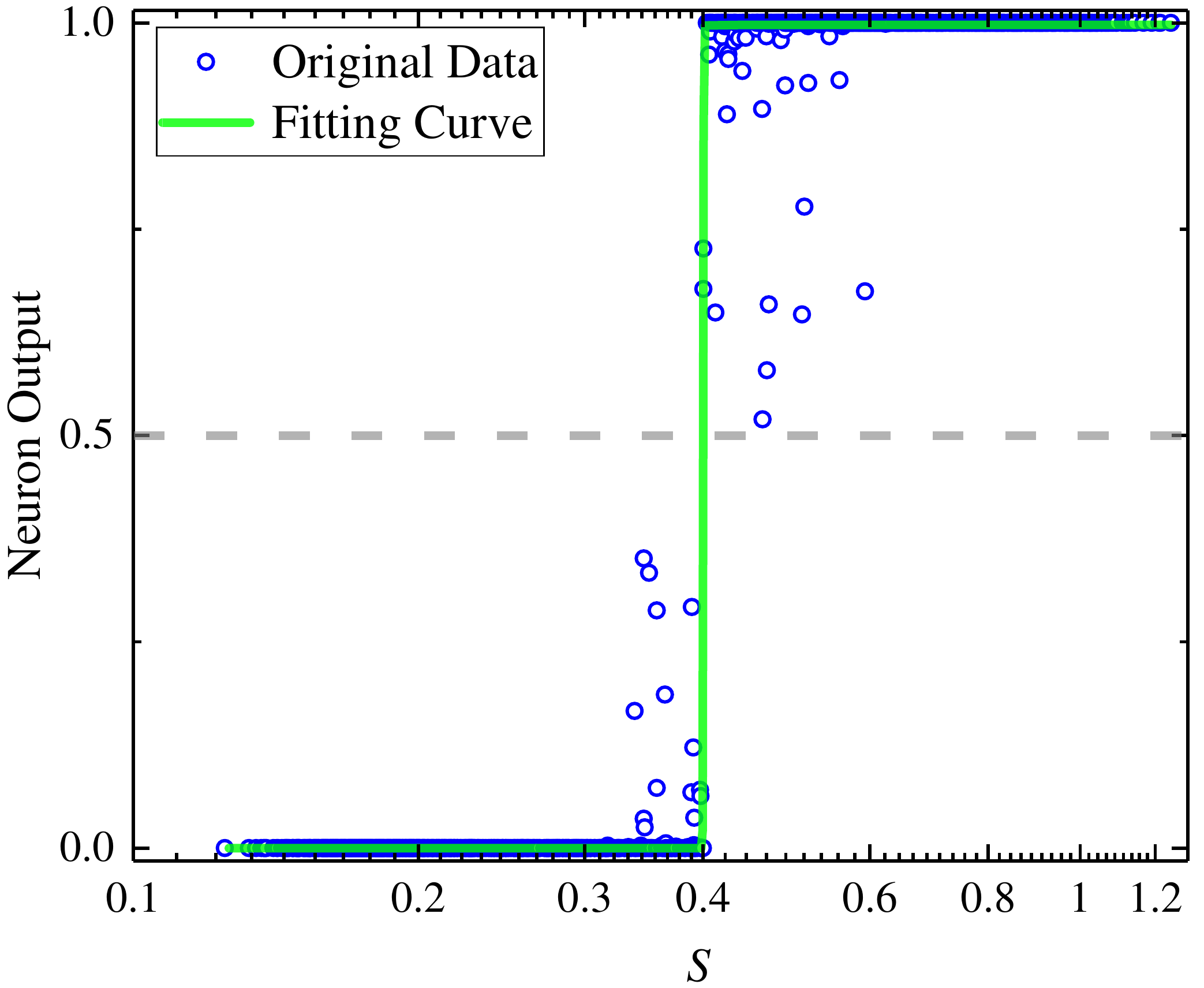}}
	\caption{\raggedright The scatter diagram of the neuron output and SDRDF.
The gray dashed line is the horizontal line with a neuron output of $0.5$.   \hfill
\label{fig:cnn_vis}}
\end{figure}
We first train the newly constructed CNN five times by the original training set and save the model parameters after each completed training.
Then we randomly choose $2520$ images of crystal and amorphous phases, importing them into the CNN, and output values of the neuron in the newly added fully-connected layer.
Note that we calculate SDRDF of each image sample before it is sent to the neural network, and we can obtain the output of the corresponding neuron after it is sent to the neural network.
We can make a scatter diagram of the neuron output and SDRDF (Fig.~\ref{fig:cnn_vis}), where each blue dot represents a image sample.
In Fig.~\ref{fig:cnn_vis} we can see that in the region where $S<0.4$ (the images are amorphous phases) the neuron output is mostly zero, while in the region where $S>0.4$ (the images are crystal) the neuron output is mostly one.
Moreover, we find a mutation from 0 to 1 in the neuron output near $S=0.4$. In summary, one can safely conclude that there is a strong correlation between the neuron output and SDRDF.

We emphasize that only the images of skyrmion crystal and amorphous phases are input to the CNN, while SDRDF is an artificially defined order parameter to distinguish between skyrmion crystal and amorphous phases.
Note also that the information about SDRDF is not directly input into the CNN.
However, the CNN learner has found out meaningful information related to SDRDF by itself.
Then we can deduce that information similar to SDRDF is also stored in this neuron, in other words, the CNN finds the order parameter of classifying crystal and amorphous phases.

Furthermore, we can predict the critical value of the spatial order parameter based on the relationship of the neuron output and SDRDF.
We use the sigmoid function to fit the relationship between the neuronal output and SDRDF. Specifically, the sigmoid function is
\begin{equation}
y(x)=\frac{1}{1+e^{-k(x-b)}} \label{fitmodel}
\end{equation}
where the parameter $k$ determines the steepness of the sigmoid function.
When $k$ tends to 0, the line of fit is horizontal, indicating that the neuron output does not correlate with the order parameter.
When $k$ tends to infinity, the line of fit is similar to a step function, indicating that the neuron output has a strong correlation with the order parameter.
The fitting curve is shown in Fig.~\ref{fig:cnn_vis}.
It can be seen that the neuron output is $1$ in the crystal part, while the neuron output is $0$ in the amorphous part.
According to the intuition about phase transitions, we can take the point where the probability is exactly $0.5$ to be the inferred critical point $S_0$ \cite{ch2017machine,carrasquilla2017machine,
liu2018discriminative,beach2018machine,dong2019machine}.
Therefore, according to the fitting curve, we can predict that the critical value of SDRDF is $S_{CNN}=0.40009$, which is very little different from the real value $S_0 = 0.40000$.
In conclusion, CNN can not only extract information of the order parameter from the original skyrmion crystal and amorphous images but also can predict the critical value of the order parameter.

\begin{figure}
	\scalebox{0.5}[0.5]{\includegraphics[width=1\textwidth]{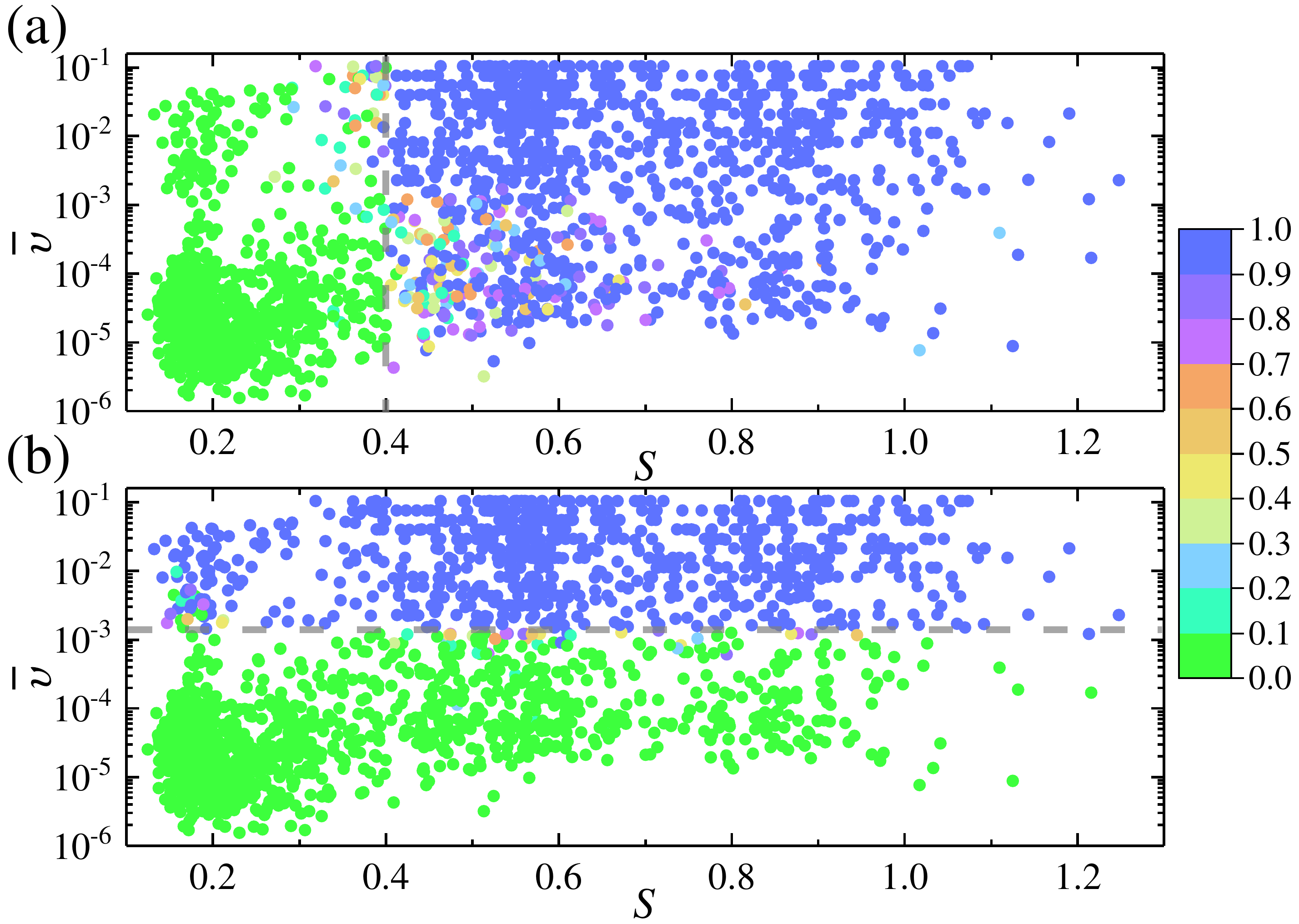}}
	\caption{\raggedright (a) the scatter diagram of the neuron one output with respect to SDRDF and mean velocity, (b) the scatter diagram of the neuron two output with respect to SDRDF and mean velocity. The gray dashed lines are the dividing lines for order parameters. Namely, the gray dashed line in (a) is $S_0=0.4$, and the gray dashed line in (b) is $\bar{v}_0=0.0014$.
\hfill
\label{fig:neuron12}}
\end{figure}
\subsection{Parameter visualization in the 3D-CNN}\label{sec.4a3}
We visualize now the neural network that identifies videos of four dynamical skyrmion phases.
Since we have demonstrated that the 3D-CNN is more successful than the CNN-LSTM at classification, we consider only the parameter visualization for the 3D-CNN.
Similar to the previous parameter visualization for the CNN, we add a fully-connected layer in front of the output layer of the 3D-CNN.
Because recognizing videos of four dynamical skyrmion phases requires both temporal and spatial information, and physically requires two order parameters (SDRDF and mean velocity), we set the number of neurons in the fully-connected layer to $2$.

After training the new 3D-CNN, the best model parameters are saved and $2520$ samples (videos of four dynamical skyrmion phases) are input into the new 3D-CNN.
Then we separately plot a scatter diagram of the output of two neurons with respect to two order parameters [see Fig.~\ref{fig:neuron12}(a) and ~\ref{fig:neuron12}(b)].
Each dot in the figure represents a sample, and the color of dots represents the values of neuron output.
It can be seen in Fig.~\ref{fig:neuron12}(a) that, when the mean velocity is constant, the output of neuron one changes from $0$ to $1$ (the color changes from green to blue) with the increase of SDRDF.
However, when SDRDF is fixed, the output of neuron one is constant no matter how the mean velocity varies, which shows that the output of neuron one is only related to SDRDF.
In other words, neuron one stores only spatial information. Similarly, from Fig.~\ref{fig:neuron12}(b), when SDRDF is fixed, the output of neuron two changes from 0 to 1 (the color changes from green to blue) as the mean velocity increases.
However, when the mean velocity is fixed, the output of neuron two is constant no matter how SDRDF varies.
It is shown that the output of neuron two is only related to the mean velocity. Namely, neuron two stores only temporal information.

\begin{figure}[b]
	\scalebox{0.5}[0.5]{\includegraphics[width=0.98\textwidth]{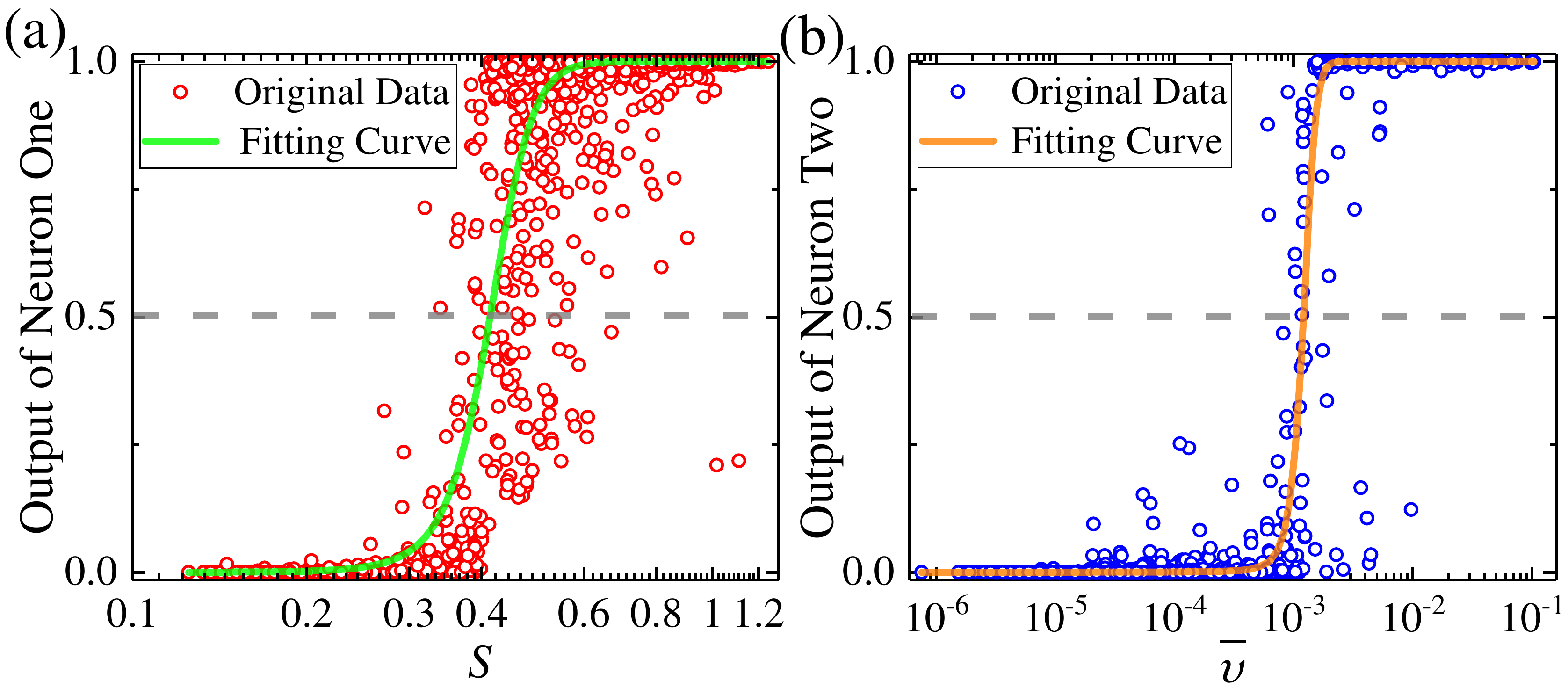}}
	\caption{\raggedright (a) The scatter diagram of the neuron one output and
SDRDF. (b) The scatter diagram of the neuron two output and mean velocity. The gray dashed line is the horizontal line with a neuron output of $0.5$.  \hfill
\label{fig:3D-CNN_vis}}
\end{figure}

We emphasize that the input of the 3D-CNN consists only in videos of four dynamical skyrmion phases, while the SDRDF and mean velocity are artificially defined order parameters to distinguish videos of four dynamical skyrmion phases, and the information of SDRDF and mean velocity is not directly entered to the 3D-CNN.
However, the 3D-CNN learner has found out meaningful information related to SDRDF and mean velocity by itself.
By visualizing the output of two neurons in the 3D-CNN we find that the output of neuron one is only related to SDRDF and the output of neuron two is only related to the mean velocity, hence one can safely conclude that neuron one stores information similar to the spatial order parameter and neuron two stores information similar to the temporal order parameter.
In other words, the 3D-CNN finds the spatial order parameter and temporal order parameter.

Moreover, we can infer the critical values of the order parameters based on the relationship between the neurons output and the order parameters.
We plot a scatter diagram of the output of neuron one versus SDRDF [Fig.~\ref{fig:3D-CNN_vis}(a)] and a scatter of the output of neuron two versus mean velocity [Fig.~\ref{fig:3D-CNN_vis}(b)], respectively.
As can be seen from the scatter (red dots) in Fig.~\ref{fig:3D-CNN_vis}(a), neuron one output has some correlation with SDRDF, and we use the sigmoid function to fit this correlation (green line).
Based on the fitting curve, when the output of neuron one is $0.5$, we can predict the corresponding critical value $S_{3D-CNN}=0.391\pm0.002$ (real value $S_0=0.400$).
Similarly, according to Fig.~\ref{fig:3D-CNN_vis}(b), we can also use the sigmoid function to fit the relationship between the output of neuron two and mean velocity, and predict the critical value of mean velocity is $\bar{v}_{3D-CNN}=0.00121$ (real value $\bar{v}_0=0.00140$).

It also needs to be added that when we perform the parameter visualization scheme, we must select the best model which has the highest accuracy on the validation set over the 5 training sessions.
The reason is that when the neural network can distinguish the input phases with maximum accuracy, the input data is compressed to its simplest and best form at the same time.
This form is closest to the physical essence, and therefore the result visualized, in this case, is the result shown in Fig.~\ref{fig:neuron12}.
Using other models for parameter visualization, we would find at this point that the neural network does not necessarily separate the temporal and spatial information, nor does it learn both the temporal and spatial order parameters (the results of these parameter visualizations are shown in Supplemental Materials).

\section{Test of parameter visualization scheme in the 3D-CNN}
\subsection{Variation of the number of neurons}
After finding that the parameter visualization scheme works surprisingly well (especially in the 3D-CNN) a related question arises:
what happens when we change the number of neurons ($N$) in the newly added fully-connected layer of the 3D-CNN?
To produce an answer to it we set the number of neurons to 1, 3, and 4 respectively, to investigate the relations between the neurons output and the order parameters.

\begin{figure}[tb]
	\scalebox{0.5}[0.5]{\includegraphics[width=0.95\textwidth]{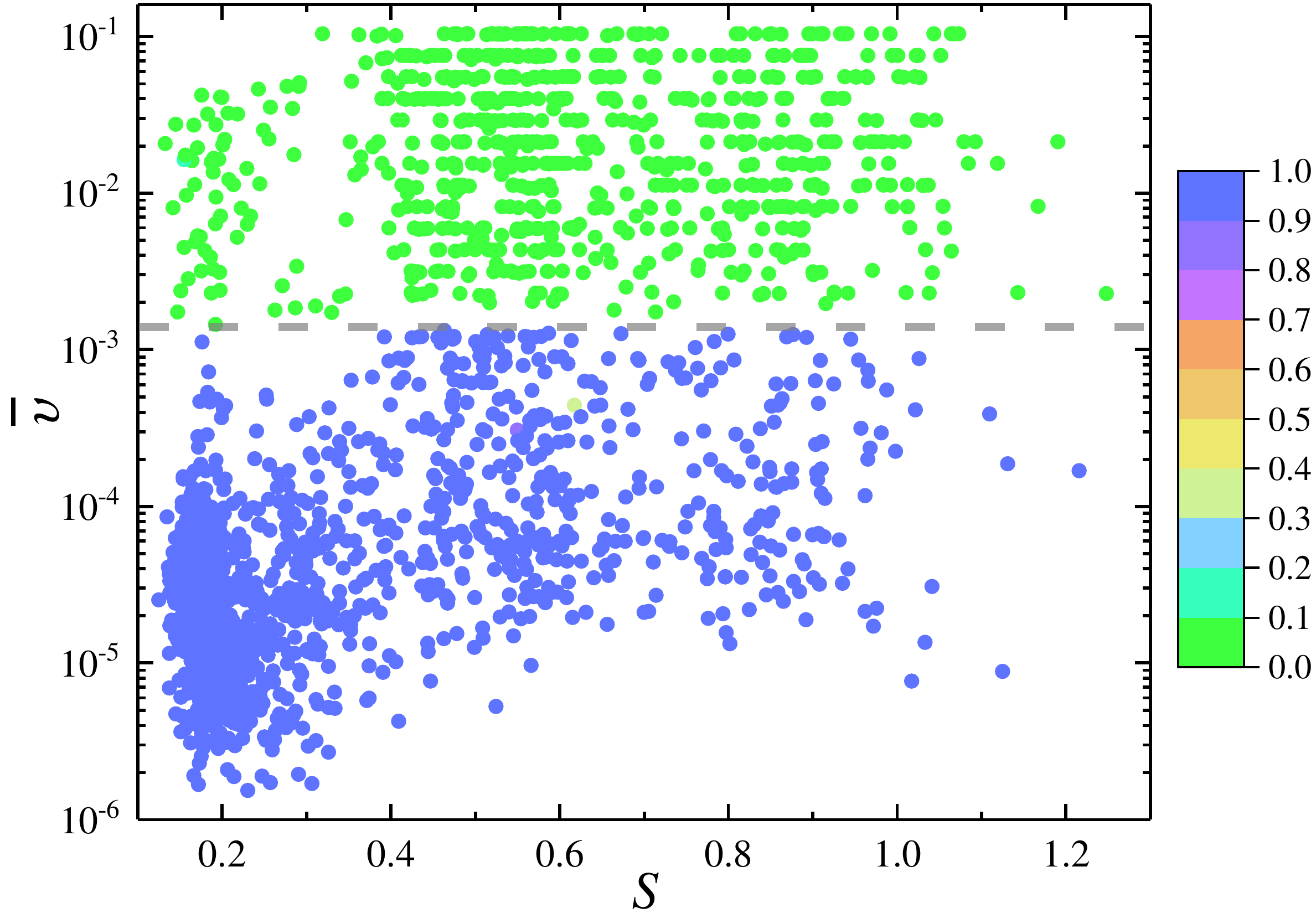}}
	\caption{\raggedright The scatter diagram of the neuron output with respect to two order parameters when the number of neurons in the newly added fully-connected layer is $1$. The gray dashed line is the dividing line for the temporal order parameter. \hfill
\label{fig:neuron1_evenly}}
\end{figure}

\subsubsection{Case $N=1$} \label{sec.5a1}
When the number of neurons (in the newly added fully-connected layer) is $1$, we use the parameter visualization scheme and draw the scatter diagram of the neuron output with respect to two order parameters (Fig.~\ref{fig:neuron1_evenly}).
We find that the neuron output is only related to mean velocity, which means that the network learns the information of the temporal order parameter.
Besides, we also find that the recognition accuracy of the 3D-CNN with one neuron is $10\%$ lower than that of the 3D-CNN with two neurons.
This suggests that using one neuron is not sufficient to distinguish videos of four dynamical skyrmion phases, at least two being needed for this.
Correspondingly, to physically identify four dynamical skyrmion phases, it is not enough to use one order parameter, at least two being required.
When we set the number of neurons in the 3D-CNN to 1 (compressing thus the information into one order parameter), the neural network would have to discard some of the information.
A related relevant question is what information would be discarded by the neural network.
The above result shows that the neural network discards spatial information and retains temporal information.
The possible reason is that learning the temporal order parameter is easier than learning the spatial order parameter.

\subsubsection{Case $N=3$ or $N=4$}  \label{sec.5a2}
\begin{figure}[tb]
\scalebox{0.5}[0.5]{\includegraphics[width=0.98\textwidth]{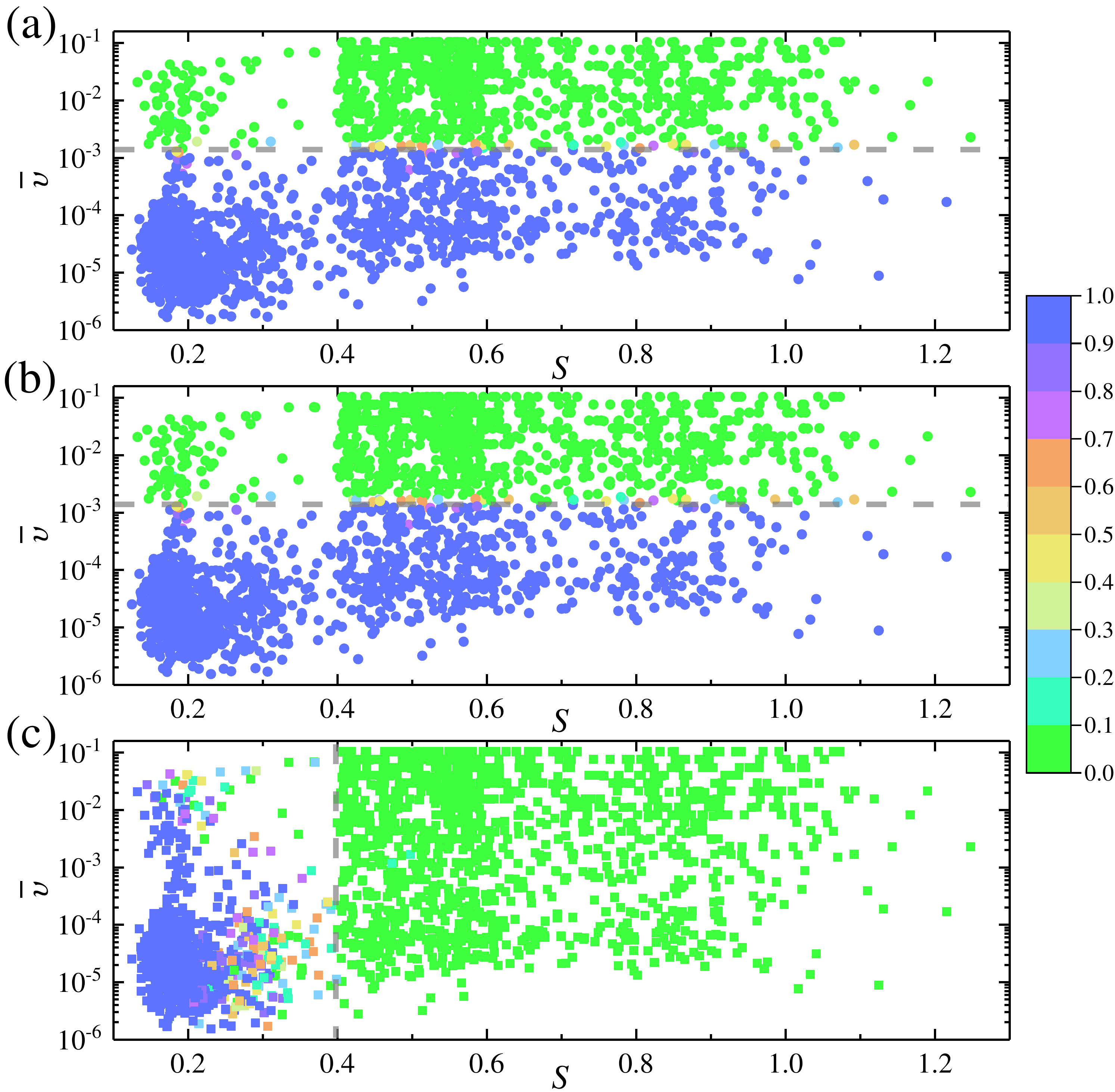}}
\caption{\raggedright When the number of neurons in the newly added fully-connected layer is $3$ (a) the scatter diagram of the neuron one output with respect to two order parameters, (b) the scatter diagram of the neuron two output with respect to two order parameters, (c) the scatter diagram of the neuron three output with respect to two order parameters.   \hfill
\label{fig:neuron3vis}}
\end{figure}

When the number of neurons is 3, we respectively plot the scatter diagrams of the neurons output with respect to two order parameters [Fig.~\ref{fig:neuron3vis}(a), \ref{fig:neuron3vis}(b), \ref{fig:neuron3vis}(c)], it can be seen that neurons one and two learn the temporal order parameter, and neuron three learns the spatial order parameter, and the output of neuron one is almost the same as that of neuron two.
This means that the information stored by one of neuron one and neuron two is redundant.
In other words, it is sufficient to use two order parameters to classify four dynamical phases of skyrmions.

When the number of neurons is 4, we respectively plot the scatter diagrams of the neurons output with respect to two order parameters (see Supplemental Materials).
We find that three neurons learn the spatial order parameter, and the other neuron learns the temporal order parameter.
In this case, it is still possible to conclude that only two order parameters are needed to classify four dynamical phases of skyrmions.

We compare the above two cases with the situation when the number of neurons is 1.
We find that when the number of neurons is 1, the 3D-CNN lacks sufficient information and thus cannot fully classify the four dynamical phases of skyrmions.
When the number of neurons is 3 or 4, our results show that the information of two neurons is enough.
Therefore, it can be determined that only two order parameters are needed to identify the four dynamical phases of skyrmions.
In conclusion, we can find out how many order parameters are needed to fully classify the input phases by our methods, which is very meaningful for exploring the unknown phases and phase transitions.

\subsection{Variation of the distribution of samples in the training set}
In the previous problems, the sample distribution in the training set we used was uniform.
In other words, the number of phases of each type in the training set is equal (for example, if the total number of the training set is 16000 with four types of phases, the number of phases of each type would be 4000).
However it is difficult to ensure that the training samples we collected in the real world are uniformly distributed, and we often deal with datasets that are unevenly distributed.
Consequently the inevitable question arises: is the neural network still able to learn the order parameters from videos when the sample is unevenly distributed?

To provide an answer to it we construct a new training set, and the distribution of samples in the training set is no longer uniform, but similar to the distribution in the phase diagram of Fig.~\ref{fig1}, which was chosen because it is closer to physical reality.
Specifically, the proportions of PG, MC, PC, and ML in the training set were $40\%$, $35\%$, $20\%$, and $5\%$, respectively, and the total number of samples in the training set remains $15960$.

\begin{figure}[tb]
\scalebox{0.5}[0.5]{\includegraphics[width=0.98\textwidth]{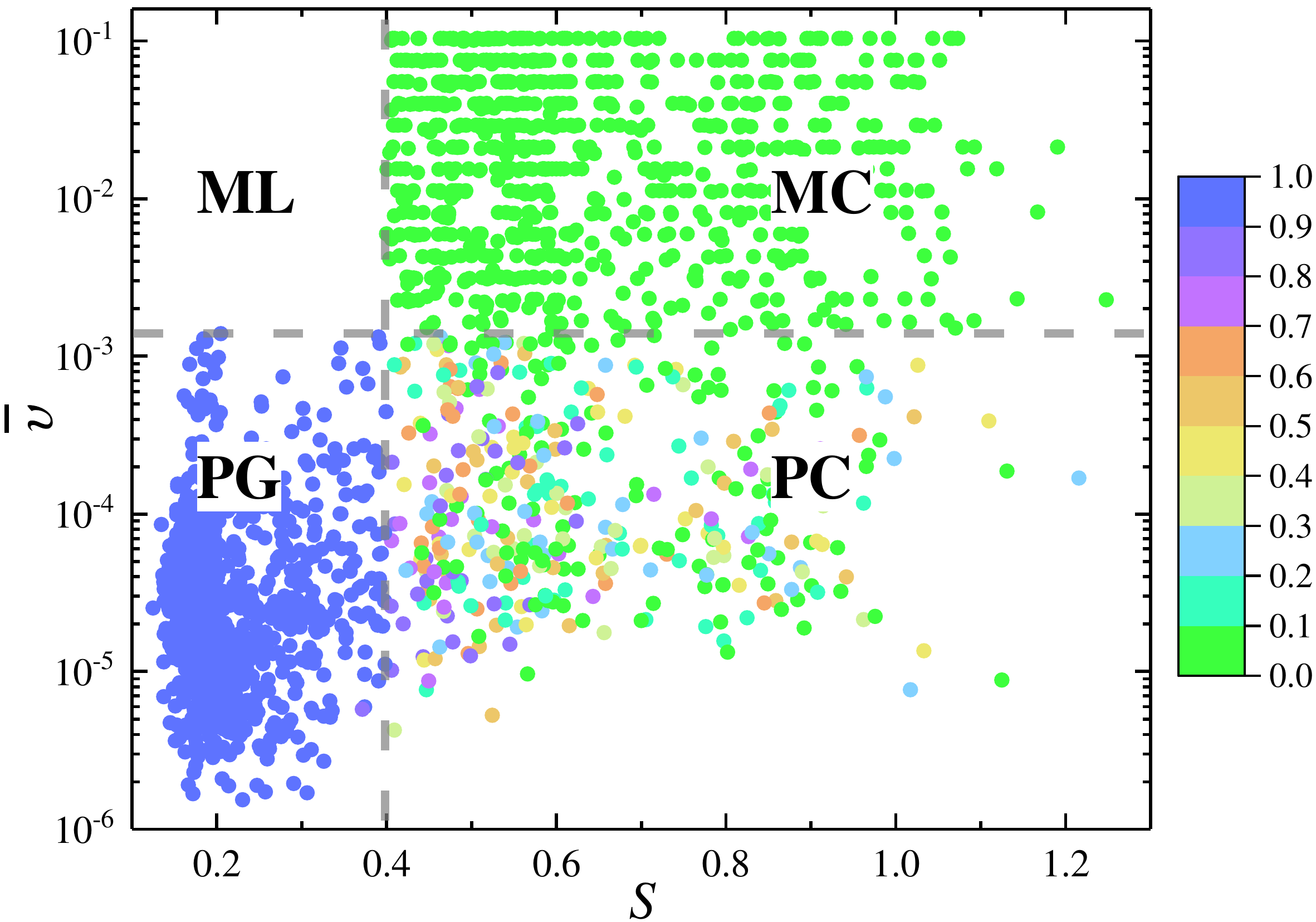}}
\caption{\raggedright The scatter diagram of the neuron output with respect to SDRDF and mean velocity when the number of neurons is $1$ and the training set is unevenly distributed.   \hfill
\label{fig:neuron1_uneven}}
\end{figure}

We first use this new training set to train the 3D-CNN which has two neurons in the newly added fully-connected layer and visualize the optimal model.
We find that the results are basically the same as those obtained in Sec.~\ref{sec.4a3}.
Two neurons learn the temporal and spatial order parameters, respectively.
We then proceed to vary the number of neurons to 1, 3, or 4 in order to explore the effect of using an unevenly distributed training set on these cases.
When the number of neurons is $1$, the scatter diagram of the neuron output with respect to two order parameters is shown in Fig.~\ref{fig:neuron1_uneven}.
We have two findings: (i) compared with the usage of two neurons, the recognition accuracy reduces by $5\%$; (ii) the ML region is empty, i.e. all ML phases are misclassified.
So when the number of neurons is 1, the 3D-CNN can only classify four types of phases into three types.
Strictly speaking, the 3D-CNN with only one neuron cannot fully recognize the four dynamical phases of skyrmions, and at least two neurons are needed.
Since at least two order parameters are needed in physics to distinguish the four dynamical phases of skyrmions, when we set the number of neurons in the 3D-CNN to 1, the neural network has to discard some information.
As can be seen from our results above, the neural network learns information about the PG, PC, and MC type phases and discards information about the ML type phase, which explains why the recognition accuracy is reduced by $5\%$ (since ML accounts for $5\%$ in the training set).
When the number of neurons is 3 or 4, we find that the results are essentially the same as those presented in Sec.~\ref{sec.5a2}. The neural network learns the temporal and spatial order parameters, and the information learned by some neurons is redundant.

In summary, when the number of neurons is 1, the neural network discards the information of the phase with the least number of samples to ensure a higher recognition accuracy, which also reflects the intelligence of the neural network.
When the number of neurons is 2, 3, or 4, the results are essentially the same as before.
This shows that the 3D-CNN trained with an unevenly distributed training set can still learn two order parameters.
Moreover, by using the parameter visualization scheme and varying the number of neurons, we can also determine how many order parameters are required to fully classify the input phases.

\section{Summary and outlook}
We develop a general video recognition method to identify the physical dynamical phases, and propose a parameter visualization scheme based on representation learning to analyze the neural network. Then we demonstrate that the neural network learns order parameters from videos.
Our method can give the number of order parameters (similar to determining the degree of freedom of a system), without requiring numerical values or the mathematical expressions of order parameters, which is a step forward in discovering physical laws from videos.

From the perspective of future applications, the techniques of video recognition can be widely used in physical experimental observations and computational simulations.
For example, employed for analyzing videos of the magnetic dynamical process \cite{seki2020propagation,deviatov2019recurrent}, it can discover important physical parameters throughout the process, and even reveal physical laws.
Or it can be used to discover some anomalies or novel physical phenomena during dealing with experimental observations of scanning tunneling microscopy.

Our method of parameter visualization comes from variational autoencoders in representation learning.
The general algorithm of variational autoencoders is to construct a network with the same input and output, and then adjust the number of neurons in the middle layer to get the minimal set of latent parameters (similar to dimensionality reduction \cite{hinton2006reducing} and data compression).
However, our method is fundamentally different from the general variational autoencoders, as the input and output are video datasets and the corresponding labels, respectively.
Therefore, our method can be seen as a combination of supervised learning and variational autoencoders, which makes it even further generalizable and capable of being applied to other supervised learning scenarios.

We note that some recent work uses variational autoencoders to extract latent parameters of partial differential equations from physical systems \cite{lu2020extracting, he2020unsupervised, wang2019emergent}, but their datasets are typically processing sequential data, which is not as complex as video data.
Moreover, they do not explicitly obtain partial differential equations.
For getting physical insights more in line with those of humans, perhaps symbolic regression \cite{udrescu2020symbolic, kim2020integration} could be used for delivering parsed expressions mapped by encoders and decoders \cite{iten2020discovering}.
If symbolic regression can be combined with video recognition techniques of supervised learning, it should be possible to obtain physical insights from videos based on mathematical expressions.

\section{Acknowledgments}
We acknowledge Wenan Guo, Xintian Wu and Huaiyang Yuan for helpful discussions about model simulations, and thank Lei Wang and Ka Shen for reading the manuscript and helpful suggestions. Also, we would like to thank Yuanyuan Mi, Zhuo Hao and Penghao Rao for technical assistance about machine learning.
This work was financially supported by the National Key Research and Development
Program of China (Grants No. 2017YFA0303300 and No. 2018YFB0407601), the National Natural Science Foundation of China (Grants No. 61774017, No. 11734004,
and No. 21421003), and NSAF (Grant No. U1930402). We also acknowledge the National Supercomputer Center in Guangzhou (Tianhe II) for providing the computing resources. The authors used Keras \cite{chollet2015keras} for machine learning.

\begin{appendix}
\section{classification problem demonstration}\label{app.A}
The aim of this appendix is to clarify the classification problem of dynamical skyrmion phases and how to make labels.
From Ref.~\cite{reichhardt2015collective}, we know that there are four dynamical phases in the system described by Eq.~\eqref{skyrmion}, which have two types of characteristics: spatial characteristics (crystal or amorphous states) and temporal characteristics (moving or pinned states).

In statistical mechanics, crystal and amorphous states can be uniquely identified by RDF. The RDF $g(r)$ was introduced in order to describe how density changes as a function of distance from a reference particle in a system of particles \cite{chandler1987introduction} and is given by
\begin{equation}
g(r)= \frac{1}{\rho_0}\frac{n(r)}{V} \label{SDRDF}
\end{equation}
where $\rho_0$ is density, $n(r)$ is the particle numbers in the ring from $r$ to $r+dr$ and $V$ is the volume.

\begin{figure}[tb]
	\scalebox{0.5}[0.5]{\includegraphics[width=0.95\textwidth]{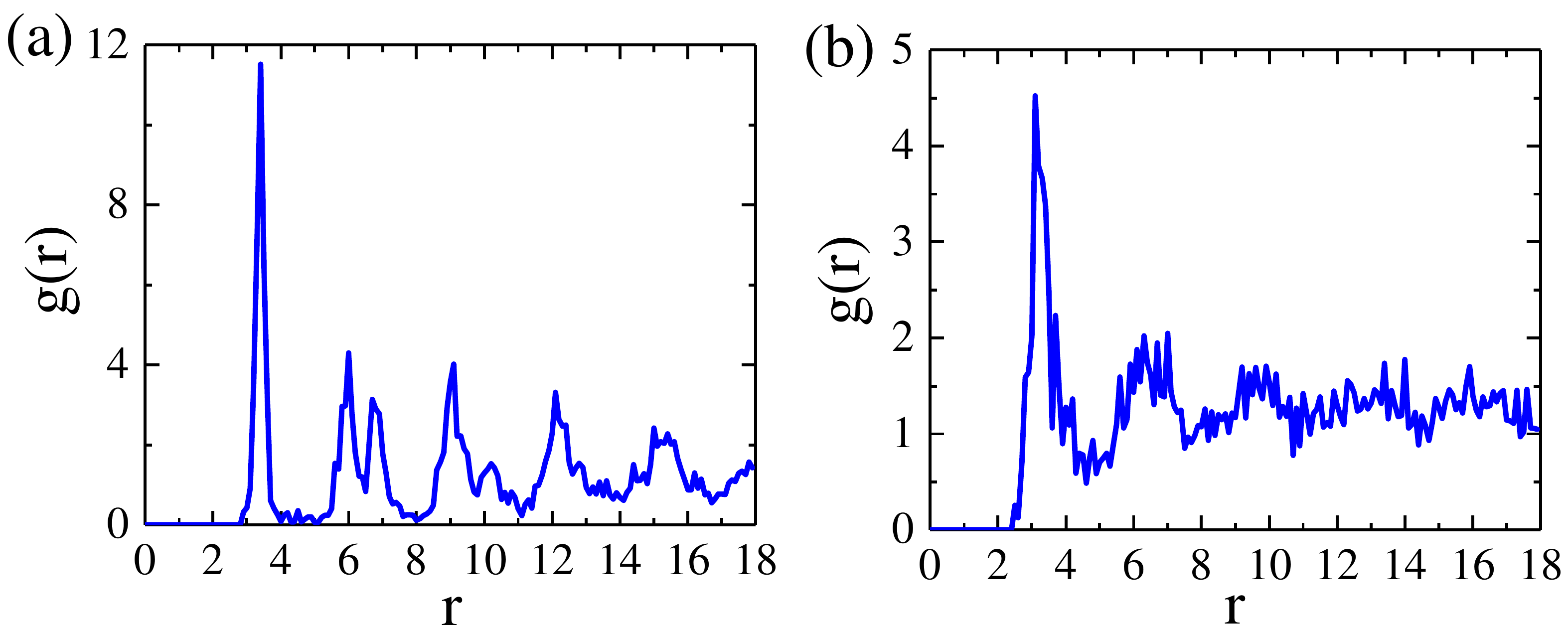}}
	\caption{\raggedright The RDF $g(r)$ as a function of the distance $r$
for crystal (a) and amorphous states (b).     \hfill
\label{fig:RDF}}
\end{figure}

Generally speaking, crystal is long-range order, and the arrangement of atoms has periodicity; noncrystal is short-range order, and the arrangement of atoms has no periodicity. The RDF of crystal and amorphous states are shown in Fig.~\ref{fig:RDF}.
The first and second peaks in RDF are the coordination numbers of the first and second near neighbour.
There are almost no resolvable peaks after the third neighbor in the RDF of amorphous state, but there are still resolvable peaks after the third neighbor in the RDF of crystal.

We calculate the standard deviation of RDF after the third peak (SDRDF) and set the critical value $S_0=0.4$ to classify crystal and amorphous states.
We assume that the weak disorder regime ($S>S_0$) is crystal while the strong disorder regime ($S\le S_0$) is amorphous state.
Therefore, when we use the CNN to classify crystal and amorphous phases in Sec.~\ref{sec.3a}, MC and PC are labeled by a label $p=(0,1)$; ML and PG are labeled by a label $p=(1,0)$.

Furthermore, we define mean-velocity ($\bar{v}$) to distinguish between moving and pinned states and set the critical value to $\bar{v}_0=0.0014$ (unit length per second) $\approx 10^{-11}$ m/s.
We assume that the low velocity regime ($\bar{v}\le \bar{v}_0$) is pinned state while the high velocity regime ($\bar{v}> \bar{v}_0$) is moving state.

In Sec.~\ref{sec.3b} and~\ref{sec.3c}, the labels of four dynamical skyrmion phases used by the CNN-LSTM and 3D-CNN are the same.
At weak disorder and low mean-velocity PC is labeled by a label $p=(1,0,0,0)$, and at weak disorder and high mean-velocity MC is labeled by a label $p=(0,1,0,0)$, and at strong disorder and low mean-velocity PG is labeled by a label $p=(0,0,1,0)$, and at strong disorder and high mean-velocity ML is labeled by a label $p=(0,0,0,1)$.

\section{details of parameter visualization scheme}\label{app.B}
We propose a scheme to visualize the parameters in the neural network and in this appendix we share some details about it.
In general, we would take the following steps (using the case when the number of neurons in the CNN is 1 as an example):
\begin{itemize}
  \item step 1. Add a newly fully-connected layer in front of the output layer of the original CNN, and set the number of neurons in the fully-connected layer to 1, getting thus a new CNN;

  \item step 2. Use the original training set to train the new CNN for five times and save the model parameters after the training is completed each time.

  \item step 3. Select the best model parameters with the highest accuracy of validation set over the five training sessions and import them into the model.

  \item step 4. 2520 samples (images of skyrmions crystal and amorphous phases) are selected and input into the new CNN, which outputs the value of the neuron in the newly added fully-connected layer and the values of the neurons in the output layer (i.e., predicted labels).

  \item step 5. Compare the predicted labels with the real labels and get accuracy. We aim to investigate what the neural network learns. Since it is impossible for the neural network to fully discriminate the input phases, we need to remove those misidentified data for which the neural network has not learned the features.

  \item step 6. Draw a scatter diagram of the neuron output with respect to the spatial order parameter.
\end{itemize}

When dealing with the problem of the 3D-CNN parameter visualization, the steps that are different from those for the CNN parameter visualization are step 1, step 4, and step 6.
In the first step, the number of neurons in the fully-connected layer is set to 2.
In the fourth step, the samples we selected are in video format. In the sixth step, we draw the scatter diagram of the outputs of the neurons with respect to two order parameters.

\end{appendix}

%

\clearpage
\renewcommand\theequation{{S-\arabic{equation}}}
\renewcommand\thetable{{S-\Roman{table}}}
\renewcommand\thefigure{{S-\arabic{figure}}}
\renewcommand\thesection{{S-\arabic{section}}}

\setcounter{section}{0}
\setcounter{table}{0}
\setcounter{figure}{0}
\setcounter{equation}{0}

\onecolumngrid
\begin{center}
  \Large
  \textbf{Supplemental Materials: Learning Order Parameters from Videos of Dynamical Phases for Skyrmions with Neural Networks}
\end{center}
\maketitle

\section{Visualization of the model parameters for each training session}
In the parameter visualization scheme, to avoid the randomness of the neural network, we trained the same neural network by the same dataset five times, and then selected the model parameters with the highest accuracy of validation set over five training sessions for visualization.
Since the order of the dataset is disrupted before each training session, this operation is equivalent to performing five times simple cross-validations.

Here we deal with the problem of parameter visualization in the same way as the previous problem of classification, for the reason that the two problems are intrinsically consistent.
The case with the highest recognition accuracy corresponds to the case where the information is compressed to its simplest and best form, i.e., closest to the physical essence.

\begin{figure}[htb]
	\includegraphics[width=0.8\textwidth]{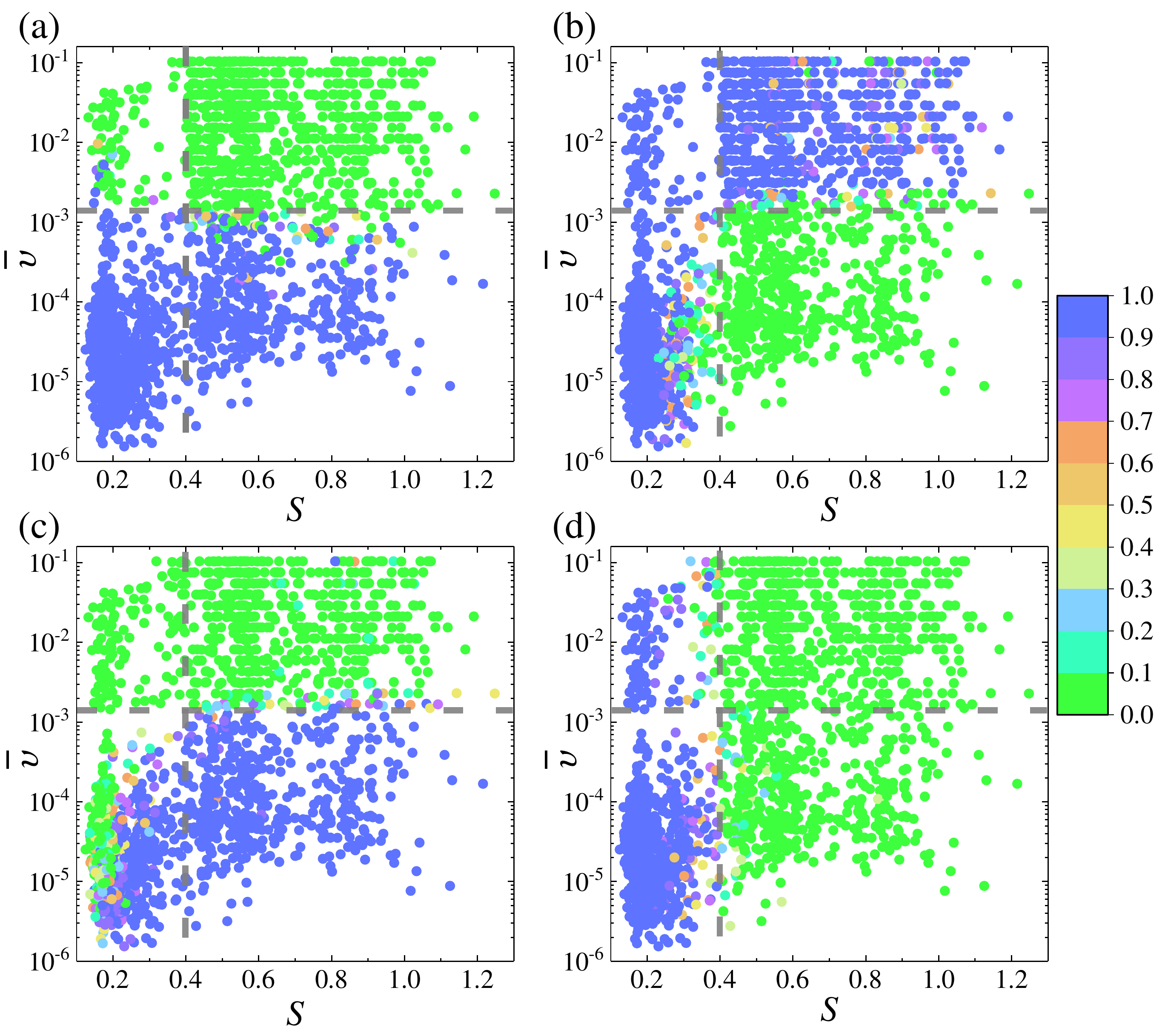}
	\caption{\raggedright Two examples of visualizing the model parameters. (a) and (b) are the results of the visualization of neurons one and two in the first example, respectively. (c) and (d) are the results of the visualization of neurons one and two in the second example, respectively.   \hfill
\label{fig:neuron2_ex}}
\end{figure}

Figure.~\ref{fig:neuron2_ex} shows examples of visualizing the model parameters derived from two of the training sessions.
The results of the first example [Fig.~\ref{fig:neuron2_ex}(a) and (b)] show that neuron one learns information about the temporal order parameter, while neuron two learns the spatiotemporal mixed information.
The results of the second example [Fig.~\ref{fig:neuron2_ex}(c) and (d)] show that neuron one learns the spatiotemporal mixed information, while neuron two learns information about the spatial order parameter.

\section{Parameter visualization of the 3D-CNN with four neurons}
When the number of neurons in the newly added fully-connected layer is $4$, we use the parameter visualization scheme and respectively plot the scatter diagrams of the neuron output with respect to two order parameters (Fig.~\ref{fig:neuron4}).
It can be seen that neuron 3 learns information about the spatial order parameter, while neurons 1, 2, and 4 learn information about the temporal order parameter.
We can conclude that only two order parameters are needed to identify the dynamical phases of skyrmions.
Besides, we also test the case where the number of neurons is 10. We find that the network can learn two order parameters, but some neurons may learn the spatiotemporal mixed information [e.g., Fig.~\ref{fig:neuron2_ex}(b) and (c)].
The reason may be that as the number of neurons increases, the information becomes more divergent and it becomes more difficult to compress the information to the best and simplest form. This makes it more difficult to learn order parameters.

\begin{figure}[htb]
	\includegraphics[width=0.8\textwidth]{neuron4.pdf}
	\caption{\raggedright When the number of neurons is 4, (a) the scatter diagram of the neuron 1 output with respect to two order parameters. (b) The scatter diagram of the neuron 2 output with respect to two order parameters. (c) The scatter diagram of the neuron 3 output with respect to two order parameters. (d) The scatter diagram of the neuron 4 output with respect to two order parameters.
\hfill
\label{fig:neuron4}}
\end{figure}

\end{document}